\documentclass[letterpaper, 10 pt, conference]{ieeeconf}  

\IEEEoverridecommandlockouts                              

\usepackage[pdftex]{graphicx}
\usepackage[noadjust]{cite}
\usepackage{algorithm}
\usepackage{algpseudocode}
\usepackage{float}
\usepackage{multirow}
\usepackage{amsmath,nccmath,amsfonts,amssymb}
\usepackage{bm}
\usepackage{ulem}
\usepackage{hyperref}
\usepackage{scalerel}

\title{\LARGE \bf
TPE-Net: Track Point Extraction and Association Network for Rail Path Proposal Generation
}

\author{Jungwon Kang$^{*,1}$, Mohammadjavad Ghorbanalivakili$^{*,2}$, Gunho Sohn$^{2}$,\\
David Beach$^{1}$ and Veronica Marin$^{1}$	
\thanks{$^{*}$These two authors are co-first authors, with equal contribution}
\thanks{$^{1}$J. Kang, D. Beach, and V. Marin are with Thales Group, Canada \newline
{\tt\small Jungwon.Kang@thalesgroup.com, David.Beach@thalesgroup.com, Veronica.Marin@thalesgroup.com}}
\thanks{$^{2}$M. Ghorbanalivakili and G. Sohn are
        with the Department of Earth and Space Science and Engineering, 
        Lassonde School of Engineering, 
        York University, 4700 Keele Street, Toronto, Ontario M3J 1P3, Canada
        {\tt\small mvakili@yorku.ca, gsohn@yorku.ca}}%
}

\begin{document}

\maketitle
\thispagestyle{empty}
\pagestyle{empty}

\begin{abstract}
One essential feature of an autonomous train is minimizing collision risks with third-party objects. To estimate the risk, the control system must identify topological information of all the rail routes ahead on which the train can possibly move, especially within merging or diverging rails. This way, the train can figure out the status of potential obstacles with respect to its route and hence, make a timely decision. Numerous studies have successfully extracted all rail tracks as a whole within forward-looking images without considering element instances. Still, some image-based methods have employed hard-coded prior knowledge of railway geometry on 3D data to associate left-right rails and generate rail route instances. However, we propose a rail path extraction pipeline in which left-right rail pixels of each rail route instance are extracted and associated through a fully convolutional encoder-decoder architecture called TPE-Net. Two different regression branches for TPE-Net are proposed to regress the locations of center points of each rail route, along with their corresponding left-right pixels. Extracted rail pixels are then spatially clustered to generate topological information of all the possible train routes (ego-paths), discarding non-ego-path ones. Experimental results on a challenging, publicly released benchmark show true-positive-pixel level average precision and recall of 0.9207 and 0.8721, respectively, at about 12 frames per second. Even though our evaluation results are not higher than the SOTA, the proposed regression pipeline performs remarkably in extracting the correspondences by looking once at the image. It generates strong rail route hypotheses without reliance on camera parameters, 3D data, and geometrical constraints.
\end{abstract}

\section{INTRODUCTION}

{
With the development of intelligent technology, the modern railway system has gradually changed from human driving mode to an autonomous and unmanned model \cite{ref_intro_1}.
One of the critical features of autonomous train systems is to avoid collision with probable third-party objects.
Such objects could be within or about to intrude on the rail routes \cite{ref_intro_6}. Through such unexpected circumstances, the autonomous train must first identify
its route among all the rail routes ahead. Then, it must precisely localize obstacles with respect to the identified route. This way, the train
can estimate the risk and react accordingly. Consequently, wayside object detection accuracy implicitly relies on the train route detection performance.
}

\begin{figure} [t]
	\centering
	\includegraphics[width=75mm]{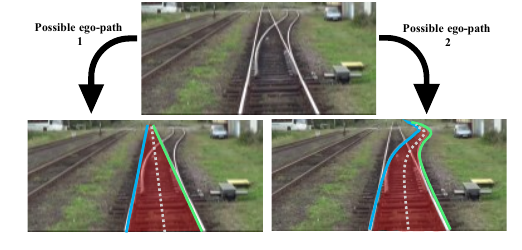}
	\caption{
The top image shows a sample input image captured by a forward-looking camera. The bottom images are the corresponding detected possible ego-paths. Other routes seen in the input image are discarded from the outcome as they are not possible ego-paths.
	}
	\label{temp_intro}
\end{figure}

{
Railway structures might incorporate switch states such as merging or diverging routes.
Thus, the train can be guided from one rail route to another.
Assuming that the switches can not
always be precisely localized, it will be more reliable to investigate all
the possible ego-paths (possible train routes) ahead rather than a single train route.}

{
In this context, a {\it track} is defined as a pair of a left rail and a right rail.
Here, we define {\it track points} as track center points located at the center between a left rail and a right rail. {\it Rail area} is also defined as the region on the ground surrounded by a track's left and right rail. Fig. \ref{temp_intro} illustrates the primary goal of this paper, i.e., identifying all the possible ego-paths along with their corresponding track points, left rail, and right rail in an input image captured by a forward-looking camera installed on a train.
}

\begin{figure*} [t]
	\centering
	\includegraphics[width=175mm]{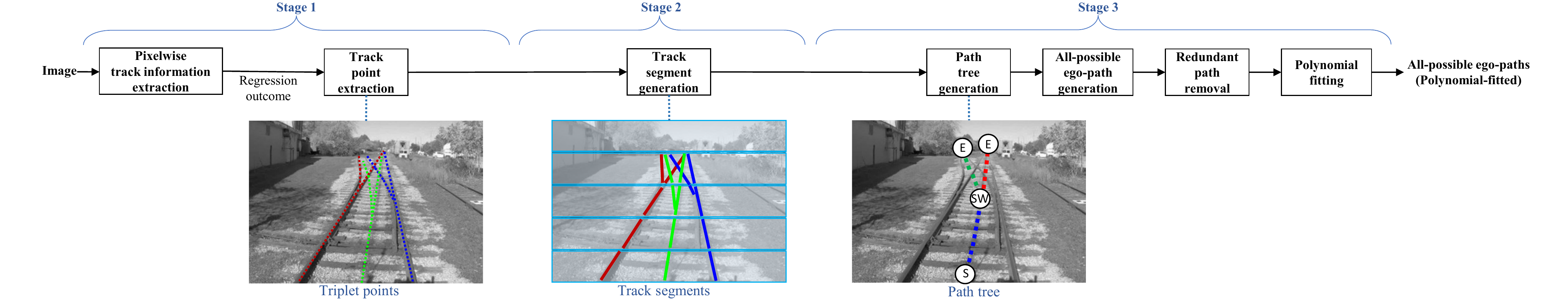}
	\caption{
	Overall diagram of our proposed rail path extraction algorithm. In the first stage, left, right, and center rail pixels are detected and associated through a fully convolutional network. Next, track segments are generated by linking the extracted track points in each sub-region. Finally, a path graph is created in the shape of a tree, covering all the possible ego-paths. In the path tree, {\it S} stands for start node, {\it SW} stands for switch node, and {\it E} stands for end node of each detected ego-path. After filtering the detected paths, polynomial fitting is performed on the extracted rail pixels to visualize extracted rails better.
	}
	\label{fig_overall_diagram}
\end{figure*}

{
In recent years, deep neural networks have shown remarkable results in image segmentation and object detection due to their high inference speed, robustness, and their ability to learn both low-level and high-level
semantic information \cite{ref_LRpaper_13}. Deep CNNs have also made huge contributions to the autonomous vehicle technology [\citen{ref_LRpaper_14},\citen{ref_LRpaper_15},\citen{ref_LRpaper_16}].
In the area of autonomous railway systems, leveraging deep neural networks in wayside obstacle detection, track extraction, switch state recognition, and track inspection has recently become a popular trend [\citen{ref_LRpaper_17},\citen{ref_LRpaper_18}]. An image-based multi-task learning network
uses Mask-RCNN with ResNet101 as its backbone to detect rail areas in the single-track case. Based on the extracted positional data, collision risk is then estimated \cite{ref_LRpaper_19}. A fully convolutional structure called DFF-Net with VGG-16 as backbone performs real-time object and rail area detection in rail scene images \cite{ref_LRpaper_20}. Rail track and rail area segmentation are also accomplished through CNNs in various studies within different single-track and multi-track scenes [\citen{ref_intro_1},\citen{ref_LRpaper_12},\citen{ref_LRpaper_18},\citen{ref_LRpaper_21},\citen{ref_LRpaper_22},\citen{ref_LRpaper_24}]. References mentioned so far have tried to extract all the rail areas or tracks as a whole, without indicating instances. Another research, however, takes one step ahead by segmenting rail tracks of complicated structures using ERFNet and then associating left-right rails to form rail pairs (track instances) using topological features \cite{ref_LRpaper_25}. Another level of railway structure understanding is also achieved through segmenting all rail areas in an image using RailCNN and then specifying all possible routes as a whole among the segmented regions \cite{ref_intro_6}.

}

{Most existing works have extracted rail elements as a whole without giving their instances. A small number of studies, however, have tried to associate left and right rail points to create track instances. This association is mainly done through the tricky RANSAC algorithm and the geometrical and topological features of the rails. Such features are thoroughly extracted from the 3D data of a LiDAR or inverse perspective transform, provided the camera parameters are available. Instead of relying on RANSAC to sample rail points, create rail hypotheses, and pick the best, we propose a deep neural network to generate strong left-right rail points hypotheses within multiple track instances, even in switch states. To find left-right rail pixel associations, the network looks once into the image without the dependency on the hard-coded prior knowledge of rail shape, topology, and geometrical constraints.
Therefore, our main contributions can be summarized as the following:
\begin{itemize}
\item
We propose a multi-task, fully convolutional neural network consisting of a segmentation and a regression branch. The regression branch estimates the degree of being a track point and pixel-level distance to the left and right rail for all the pixels in the input image. This way, each track instance's center, left, and right rail pixels are extracted efficiently.
\item
Through a bottom-up process, we construct a path tree containing topological and geometrical features of all the possible ego-paths in front of the train, which is done by spatially clustering rail pixels extracted by the neural network.
\item
We validate our proposed algorithm using the RailSem19 public benchmark, containing challenging scenes of various ambient conditions and rail structures.
\end{itemize}
}

\begin{figure} [t]
	\centering
	\includegraphics[width=85mm]{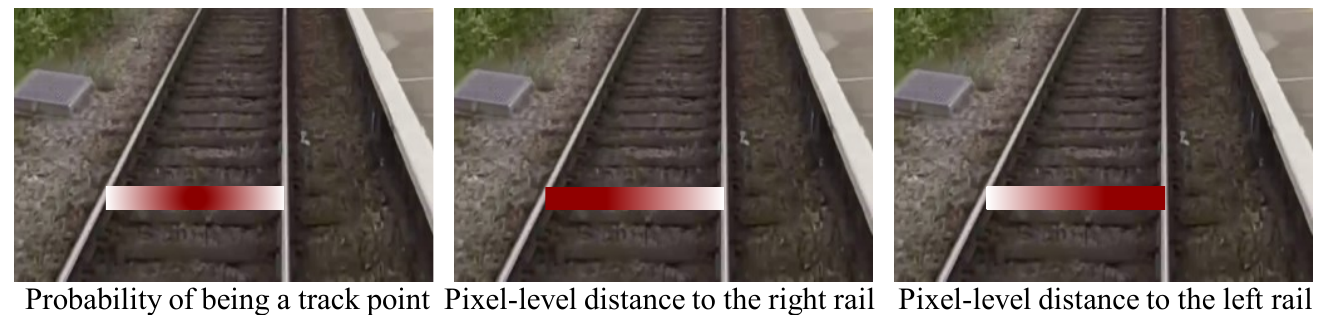}
	\caption{
Outputs of the first version of the network for its different regression tasks within all pixels of the input image. The outputs are all in the form of 1-channel heatmaps. Here, the heatmap value gets larger as the color goes darker.
	}
	\label{fig_cen_left_right}
\end{figure}

\section{Methodology}

\subsection{System Overview}

{The overall structure of the proposed system is described in Fig. \ref{fig_overall_diagram}.
To tackle complex structures of paths,
the proposed system was designed to have a bottom-up process \cite{ref_LR_3}, which was realized through the following three major stages:
In the first stage, pixel-level target points (left, right, and center rail pixels, called {\it triplets}) are extracted from the input image.
In the following stage, track segments are generated from the track points.
Finally, linking the track segments generates all the possible ego-path trajectories.
}
%

\subsection{Pixel-level Track Point Extraction}
{
Inspired by reference \cite{ref_methodology_4}, we model each track instance as a group of corresponding track points.
To extract the track points,
we designed a deep neural network called {\it Track Point Extraction Network (TPE-Net)} that regresses the degree (probability) of being a track point for all the pixels through a heatmap.
Because it is also critical to identify left and right rail pixels for a track point,
our deep neural network is supposed to
regress pixel-level horizontal distance to the track point's left and right rail within all pixels. Each distance can be estimated through a heatmap as well. Fig. \ref{fig_cen_left_right} illustrates an overview of the expected outputs of our network.
}

\begin{figure} [t]
	\centering
	\includegraphics[width=70mm]{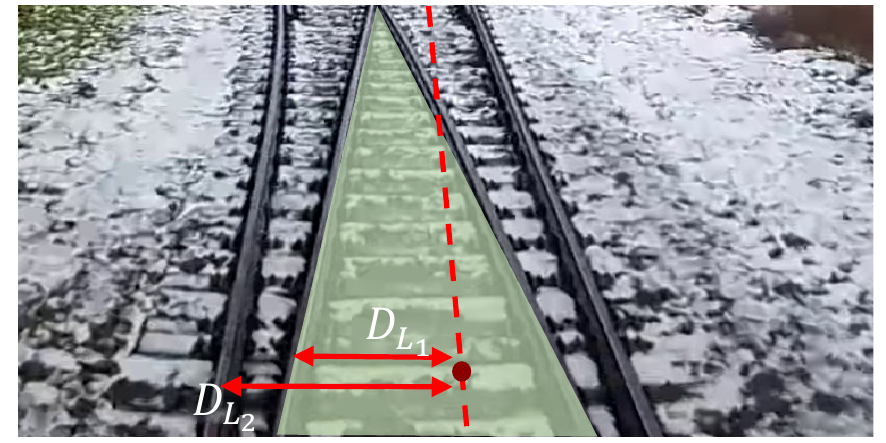}
	\caption{
In the switch region, pixels inside the shared rail area (annotated in green) have more than one left/right distance. Here, the extracted track point shown by a red dot corresponds to the rightmost rail track.
	}
	\label{fig_left_right_issue}
\end{figure}

\begin{figure*} [t]
	\centering
	\includegraphics[width=170mm]{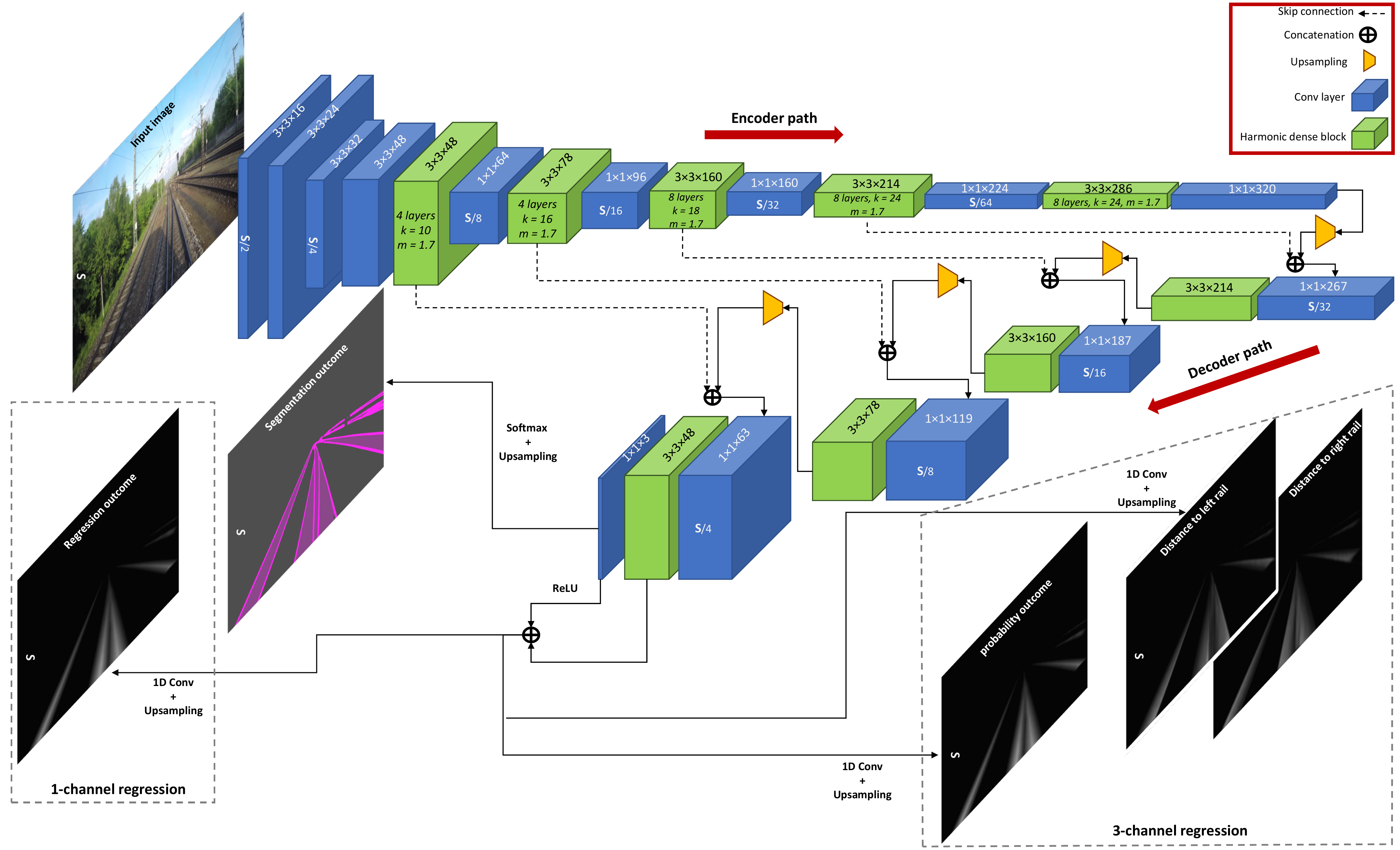}
	\caption{
A detailed structure of the proposed fully convolutional TPE-Net that outputs segmentation and triplet coordinates' regression within 2D images of rail scene. There are two different designs for the regression branch in our proposed network.
	}
	\label{temp_fig_tpe_net}
\end{figure*}

{Coordinates of the peaks within the probability heatmap are found through non-maximum suppression, resulting in the track points' locations. Then, the distance to the left and right rail pixel for each extracted track point equals the value of left/right distance heatmaps in the corresponding track point location. Here, triplet point extraction might be inaccurate in rail switches. As illustrated in Fig. \ref{fig_left_right_issue}, there is more than one left/right distance associated with each pixel in the shared region, making it impossible to define 1-channel ground truth left/right heatmaps. Therefore, regressed distances will be unreliable if the track points lie within the shared rail area. We call our first version of the regression method {\it 3-channel} regression (i.e., probability, left, and right distance). To overcome the 3-channel regression issue, we have come up with the latest version of TPE-Net, in which the probability heatmap is modified so that it regresses two types of attributes at the same time, i.e., the degree of being a track point and the equal horizontal distance to the left and right rail, for all the estimated track points. In other words, coordinates of the peaks in the 1-channel heatmap give the track point locations, while the peak values estimate the equal distance to the left and right rails. The resulting heatmap will be further explained in section \ref{section_loss}, where we define the ground truths and our proposed loss function. Throughout the following sections, we call our modified version of regression as {\it 1-channel} regression.
}

{
Inspired by the great success of recent multi-task networks \cite{ref_methodology_1} that perform semantic segmentation and other auxiliary tasks, we use multi-task learning to leverage semantic segmentation to produce desired regression outcomes. This way, a shared representation of semantic and geometric features is obtained.
Our network is based on
the semantic segmentation network \cite{ref_methodology_2} extended by \cite{ref_methodology_3}.
A detailed illustration of the architecture is shown in Fig. \ref{temp_fig_tpe_net}. Our multi-task network has a shared backbone encoder as HarDNets that have achieved an acceptable segmentation accuracy at a considerably lower inference time compared with DenseNet and substantially deeper ResNets \cite{ref_methodology_3}. There are also two decoder branches for performing two different tasks. One decoder branch performs semantic segmentation \cite{ref_methodology_2},
while the other branch was designed as a regression module \cite{ref_methodology_4} to extract triplets. Two different designs are associated with the regression branch, i.e., the 3-channel regression branch and the 1-channel regression branch.
}

{
Our decoder path is somewhat the mirrored version of the encoder path, containing four HarDblocks with the same design as the fourth, third, second, and first encoder HarDBlock. To recover full resolution, skip connections and four transition-ups in the form of bilinear interpolation are employed. Upsampled feature layers are concatenated with the feature layers transferred through skip connections. The number of input channels to each decoder HarDBlock is divided by two using 1D convolutions to overcome the sudden rise in memory demand due to the upsampling and concatenation \cite{ref_methodology_2}. The output of the last decoder HarDBlock is fed into separate 1D convolutions through two different branches, representing the segmentation branch and regression branch. Note that the regression branch fuses segmentation feature layers with the output of the last HarDblock to leverage semantic information in predicting a more accurate heatmap. Here, all the convolution layers have the same structure of Conv + BN + ReLU, except for the segmentation output which Softmax predicts.
}

\subsection{Track Segment Generation}

{Here, track segments are generated from the extracted track points.
To this end, the region for the input image $ I $ is first divided into non-overlapping sub-regions,
where $ i $th sub-region $ I_i $ has the same width $ W $ as $ I $, and a small height $ h $.
The process of generating track segments from the triplets is illustrated in Fig. \ref{fig_overall_diagram}.
In each sub-region, spatially neighboring track points are clustered to generate track segments.
The clustering is done by checking track points at each row of the sub-region,
in the direction from the bottom-most sub-region to the top-most sub-region, i.e., close to far direction.
At each row, track segments generated at previous rows are associated with a track point in the current row.
If a track segment is spatially close to the track point,
the track segment is augmented with the track point.
Otherwise, i.e., if the track point is spatially far from all the existing track segments,
a new track segment is generated with the track point.
Through the above steps, we generate track segments $ {\bf{S}} $ for all the sub-regions,
where $ {\bf{s}}_{i} = \{ {\bf{s}}_1^i, ... , {\bf{s}}_{n_s}^i\} $ in $ {\bf{S}} $ denotes track segments in $ i $th sub-region.
Here, a track segment $ {\bf{s}}_j^i $ is a set of track points.
}

\subsection{All-possible Ego-path Generation}

{The final step is to generate complete paths from the track segments $ {\bf{S}} $. 
To this end, we construct a rooted tree $ {\mathcal{G}} $ that includes geometric and topological information about the paths. 
The tree $ {\mathcal{G}} $, dubbed {\it path tree} here, consists of nodes $ {\mathcal{N}} $ and edges $ {\mathcal{E}} $. 
The nodes $ {\mathcal{N}} $ represent topologically meaningful locations, which include three types of nodes: 
(i) start node, which indicates a point where a path starts, 
(ii) end node, which indicates a point where a path ends, and 
(iii) switch node, which indicates a diverging point in a path.
The edges $ {\mathcal{E}} $ represent path trajectories between two nodes.
}

{
The path tree $ {\mathcal{G}} $ is constructed by creating nodes and edges from track segments $ {\bf{S}} $, as illustrated through Fig. \ref{fig_overall_diagram} .
The start node is supposed to exist below the bottom middle of the image $I$.
Construction of $ {\mathcal{G}} $ is done by handling track segments
from the bottom-most sub-region to the top-most one, i.e., close to far direction.
The clustering of spatially neighboring track segments over two neighboring windows starts from a track segment located around the center in the bottom window as a start node.
At the beginning of the construction,
track segments in the bottom-most sub-region are associated with the start node based on the spatial distance.
Next, the spatially close track segments are used to create edges started from the start node.
Then, each track segment is associated with existing edges, and the track segment is merged into the spatially close existing edge.
This association process is done in the direction from the bottom to the top of the image.
If two or more track segments under the current association check are spatially close to the same existing edge,
a switch node is created at the end of the existing edge,
and new edges are created from the track segments.
Note that this switch node is considered a junction of two different paths, i.e., a diverging point in a switch region.
With the constructed $ {\mathcal{G}} $, all the possible ego-paths are obtained by simply traversing from end nodes to a start node.
Each possible ego-path is a path trajectory consisting of center points and corresponding left and right rail points for the path.
}

{On top of exploiting a semantic segmentation network to regress triplet points locations, we use the segmentation output of our proposed network through the post-processing stage to reduce regression errors. If the regressed left-right rail pixel does not lie within a specific horizontal distance to a rail class pixel, we shift the estimated location to its nearest rail class pixel with the highest probability in each image row. The proposed error compensation is performed on the condition that the rail area width decreases in a close to far distance direction due to the perspective effect. Finally, a polynomial is fitted to each left and right rail of a proposed path through least squares curve fitting to obtain a solid visualization of the extracted rails.
}

\subsection{Loss Function} \label{section_loss}
{
The dataset chosen to train the network and evaluate our rail path extraction algorithm is RailSem19 \cite{ref_result_1}, a segmented dataset with 8500 images covering categories such as rail track and rail area. Also, coordinates of all left and right rail pixels for each track instance are provided.}

{For the segmentation task, we use the same ground truth masks provided in the dataset, except that the number of classes can be reduced from 19 to 3, i.e., rail track, rail area, and background. However, we go through the following steps to produce ground truth data for the regression task. In the case of 3-channel regression, probability heatmap ground truth within each rail area row increases linearly from 0 to 1, from each left/right rail pixel to the corresponding track point. In shared rail areas, pixels take the maximum among the corresponding probabilities associated with each track instance. Distance heatmaps within each rail area row increase from 0 to rail area width, starting from the left/right rail pixel. Here, pixels of shared rail areas are given values based on the order of track instances, which will be troublesome.
In the case of 1-channel regression branch, we define a 1-channel heatmap $I_{GT}$ to serve as our ground truth. Each rail area pixel value equals the minimum of horizontal distance to the left rail $d_L$ and right rail $d_R$. In the case of overlapping rails, pixels in the shared region take the maximum among the corresponding distances associated with each track instance. So, the ground truth heatmap for the regression task is created according to (\ref{eq_1}), in which {\it n} is the total number of rail areas covering pixel {\it (x,y)}. As we only deal with the peaks (corresponding to centers) in the regression heatmap, and the peaks never overlap even in shared rail areas, we do not face the issues associated with the 3-channel regression branch.
\begin{equation} \label{eq_1}
I_{GT}(x,y) = {\bf Max}(min(d_{L_1},d_{R_1}),...,min(d_{L_n},d_{R_n}))
\end{equation}
}

{We define the multi-task loss as a weighted sum of bootstrapped cross entropy (BCE) \cite{ref_result_2} loss for segmentation and L1 loss for regressions.
BCE loss for segmentation ($Loss_{seg}$) through each image of the batch is defined as (\ref{eq_2}). Here, $\it W$ and $\it H$ are the image width and height, respectively, $\it ind$ stands for indicator function, $\it Loss_{seg}(x,y)$ is each pixel's cross-entropy, and $t_{K}$ is the highest possible threshold such that over all the $W\times H$ pixels, at least $\it K$ pixels have the indicator output of 1.
\begin{equation} \label{eq_2}
\scaleto{Loss_{seg} = \frac{1}{K}\sum_{x=1}^{W} \sum_{y=1}^{H} ind(Loss_{seg}(x,y)>t_K)\times Loss_{seg}(x,y)}{30pt}
\end{equation} 
Likewise, each regression loss for a heatmap of the batch ($Loss_{reg}$) is obtained through (\ref{eq_3}). Here, $I_{est}$ is a regression output of TPE-Net, and $\it I_{G}$ is the ground truth heatmap. Finally, total loss ($Loss$) through each batch is defined using (\ref{eq_4}) in the case of 1-channel regression, and (\ref{eq_5}) in the case of 3-channel regression. Here, $\it BS$ stands for the specified batch size.
\begin{equation} \label{eq_3}
Loss_{reg} = \frac{1}{W\times H} \sum_{x=1}^{W} \sum_{y=1}^{H} abs(I_{G}(x,y) -  I_{est}(x,y))
\end{equation}
\begin{equation} \label{eq_4}
Loss = \frac{1}{BS} \sum_{i=1}^{BS} 0.4\times Loss_{reg}^i + Loss_{seg}^i 
\end{equation}
\begin{equation} \label{eq_5}
\begin{split}
Loss & = \frac{1}{BS} \sum_{i=1}^{BS} 0.2\times Loss_{reg(distances)}^i \\ 
&  + 20\times Loss_{reg(probability)}^i + Loss_{seg}^i
\end{split}
\end{equation}
}

\section{Experimental Results}
\subsection{Training}
{
We train the network using the stochastic gradient descent optimizer on 6000 training and 1000 validation images. The images are resized from the original resolution of $1080\times1920$ to $540\times960$. The initial learning rate of the network is 0.001 with a polynomial decay scheduler, the momentum parameter is 0.9,
the weight penalty is 0.0005, the batch size is 8, and in the case of 1-channel regression with three segmentation classes, the number of epochs is 100. In addition, the loss threshold for BCE ($t_K$) is set to 0.3, and the minimum number of pixels associated with BCE loss ($K$) is set to 8192.
We train the network on a computer with NVIDIA GeForce RTX 3090 GPU and Pytorch 1.10.2, taking around 48 hours to finish the training process.
}


\begin{table*} [t]
	\centering
	\caption{
	Performance of TPE-Net rail path extraction on 1500 randomly selected images of RailSem19 dataset
	}
	\includegraphics[width=175mm]{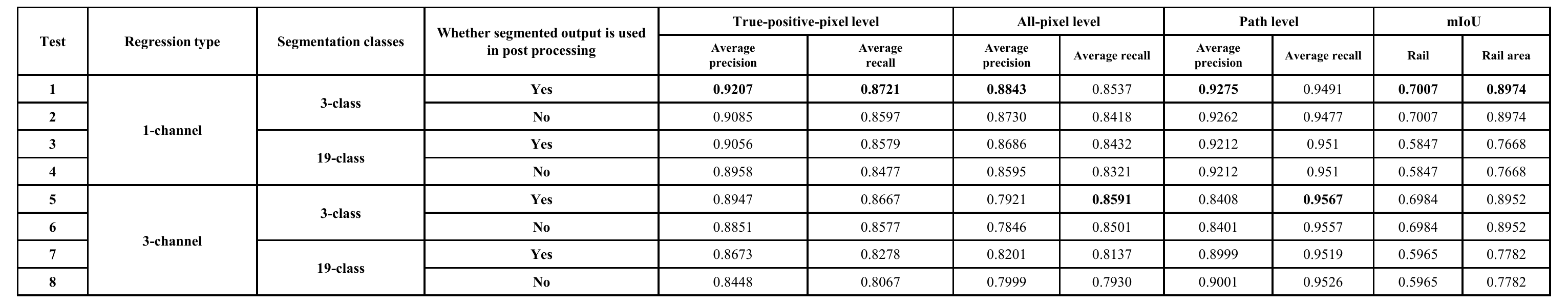}
	\label{table_result}
\end{table*}

\subsection{Results and Discussion}
{We employ the mean Intersection over Union (mIoU) criterion to evaluate segmentation performance. However, we need to modify the RailSem19 dataset to cover only triplets of the possible ego-paths to evaluate our path extraction. As previously seen, TPE-net outputs triplets of all the rails routes within an input image. Then, our proposed post-processing generates a path tree containing only the possible ego-paths. Therefore, if we feed the ground truth rail coordinates of the RailSem19 test set to our post-processing, ground truth data of the possible ego-paths will be generated.
}

{We evaluate our path proposal generation algorithm in three levels: true-positive-rail pixel level, all-rail pixel level, and rail path level \cite{ref_LRpaper_25}. To match the ground truth paths with the estimated ones, we first explain how to measure the F1 value between the two to serve as the matching rate based on which we can judge correspondences. Accordingly, we first pick the coordinates of each ground truth left/right rail pixel. If an estimated rail pixel exists in a defined vicinity of the ground truth coordinates, two pixels are matched, and the ground truth is considered a true-positive (TP). Ground truths without matching estimated pixels form false-negatives (FN) while remaining estimated pixels with no ground truth matches are false-positives (FP). With TPs, FPs, and FNs in hand, the F1 measure is calculated. Here, ground truth and extracted paths exceeding a minimum matching rate are matched one by one.
}

{If unmatched ground truth and estimated paths are discarded, true-positive-rail pixel level precision and recall are calculated among the matching paths, as explained above. Here, if FPs and FNs of the unmatched estimated and ground truth paths are also considered, all-pixel level precision and recall are measured. We expand our view from the single pixels to the whole path for the path level performance metrics. Therefore, matching paths provide TPs, unmatched ground truths form FNs, and the unmatched extracted paths are FPs.
}

{
Numerical results of TPE-Net rail path extraction based on the defined performance metrics are provided in Table \ref{table_result}. If we reduce the original 19 segmentation classes of RailSem19 to 3, exploit segmentation output in our post-processing, and rely on 1-channel regression, the highest average precision and recall in most levels are obtained. Here, network GPU time is 0.0156, while post-processing CPU time on AMD Ryzen 9 5900X 12-core processor will be 0.0727 seconds per image.
}

\begin{figure} [t]
	\centering
	\includegraphics[width=85mm]{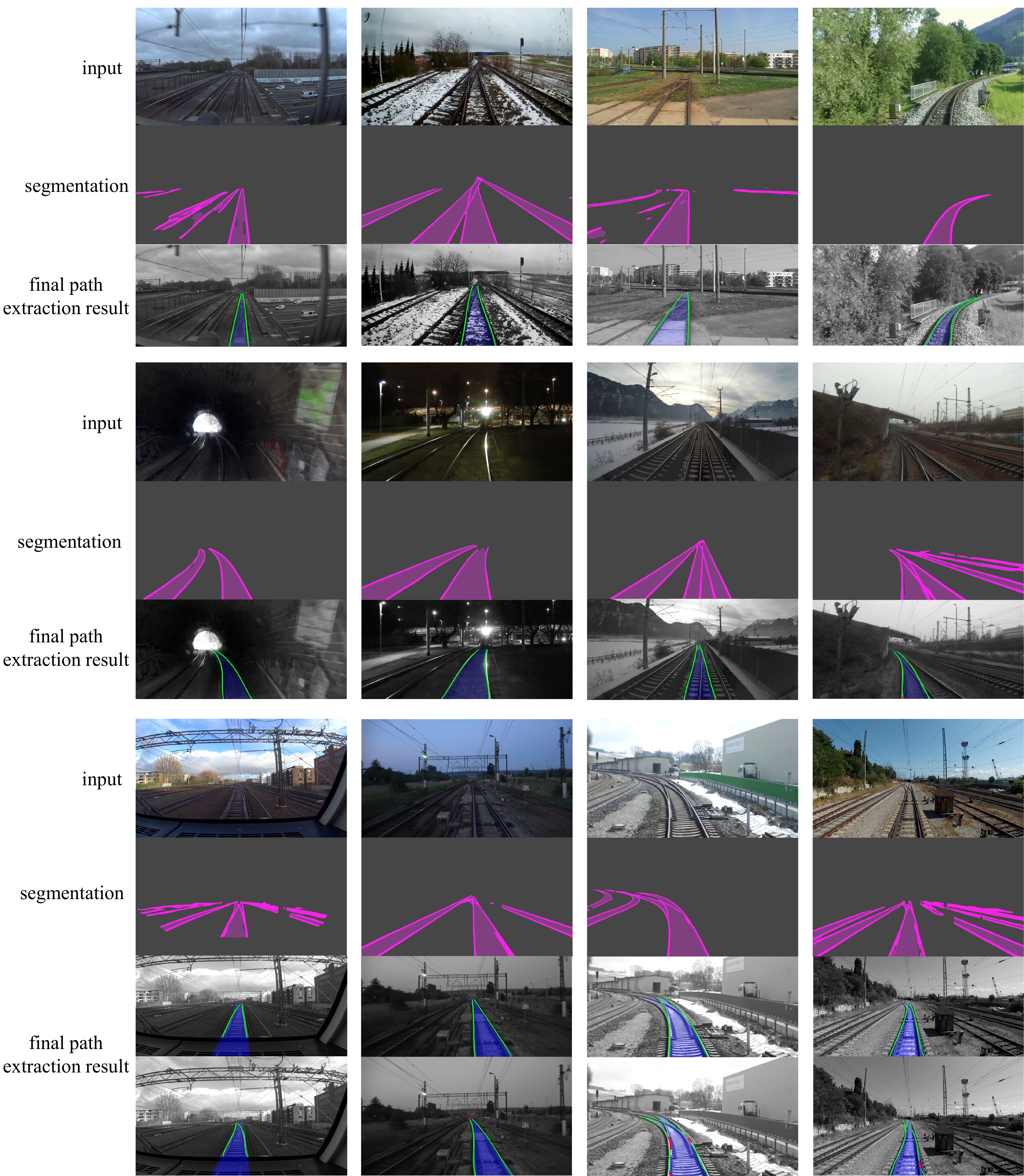}
	\caption{
Visual results of the proposed path extraction algorithm on some sample test images of the RailSem19 dataset. The results include the input image, segmented output of the TPE-Net, and the final rail path extraction result. Input images are selected to cover multiple kinds of weather, illuminations, and switch state conditions. In the final path extraction result, the ground truth rail area is annotated in blue, TP pixels are green, and FP pixels are red. Also, each extracted path is shown in a separate image for clear visualization. 
	}
	\label{TP_ result_visualization}
\end{figure}

{
Fig. \ref{TP_ result_visualization} illustrates some path extraction results on RailSem19 test images using 1-channel regression. The results include the input image, segmented output, and path extraction outcome. In the images showing the outcome, the ground truth rail area is annotated in blue, TP pixels are green, and FP pixels are red. Also, each detected path is drawn in a separate image for better visualization.
}

{Reference \cite{ref_LRpaper_25}, assumed as SOTA, is one of the few papers that has reached remarkable results in track segmentation and left-right rail association to extract all the path instances. With 95.4\% true-positive-rail pixel level, 94.87\% all-pixel level, and 98.18\% path level average precision, their proposed algorithm outperforms ours. However, TPE-net rail path extraction results are comparably promising due to the following points:
(1) SOTA paper tests their proposed algorithm on a private dataset. However, we have calculated the performance metrics using a challenging, publicly released benchmark, trying to achieve the highest possible generalization.
(2) In the SOTA paper benchmark, rail scene images are all captured using one single camera installed on a specific train. Thus, camera parameters are used to extract inverse perspective transform. Consequently, extracted 3D positional data provides useful topological information for rail detection and association. However, such a feature is not feasible when dealing with RailSem19, as the images are taken in various regions worldwide.
(3) SOTA paper relies heavily on post-processing to associate left-right rails, while no result on the inference or run time is provided. However, our proposed TPE-Net is capable of left-right rail pixel detection and association through its regression branch at 64 frames per second without dependency on hard-coded prior knowledge of railway structure.
}

\section{Conclusions}
{ 
This study proposed an image-based rail path extraction algorithm for autonomous trains based on a fully convolutional neural network.
At first, the multi-task network segments the input image into rails and other classes while regressing the rail pixels of each track instance through either regression branch designs. Second, track segments are generated through spatially clustering the extracted track points. Finally, a path tree containing topological features of the paths is generated through spatially clustering track segments.
}

{
Experimental results on the RailSem19 dataset show that if we reduce segmentation classes from 19 to 3, use the segmentation outcome in the post-processing to compensate for the regression errors, and prefer 1-channel regression over 3-channel regression, we reach the true-positive-pixel level average precision and recall of 0.9207 and 0.8721, all-pixel level average precision and recall of 0.8843 and 0.8537, and path level average precision and recall of 0.9275 and 0.9491, respectively.
}

{
Even though higher performance metrics of the same task on different private benchmarks exist among previous studies, our results are acceptable as we have associated the left-right rail points through an end-to-end trainable network with real-time performance instead of relying heavily on post-processing and prior topological knowledge. Also, we have shown interest in detecting only the possible ego-paths for the sake of autonomous trains. However, our proposed network occasionally performs unreliably in switch regions, estimating high center point probability within the shared areas by judging based on rail track salience. Therefore, designing a more robust rail pixel detection pipeline will be part of our future work.
}

\end{document}